%% file: FedBWOppr.tex
\documentclass[10pt,conference]{IEEEtran}

\usepackage{fancyhdr}

\IEEEoverridecommandlockouts
\usepackage{amsmath}
\usepackage{subfig}
\usepackage{comment}
\usepackage[noend]{algpseudocode}
\usepackage[T1]{fontenc}
\usepackage{lmodern}
\usepackage{multirow}
\usepackage{amsmath}
\usepackage{mathtools}
\usepackage{xcolor}
\usepackage[ruled, vlined, linesnumbered]{algorithm2e}
\usepackage{algcompatible}
\usepackage{amssymb}
\usepackage{bbding}
\usepackage{pifont}
\usepackage{tikz}

\usepackage[compact]{titlesec}         

\usepackage[rawfloats=true]{floatrow} 
\restylefloat{figure}     
\ifCLASSOPTIONcompsoc
  \usepackage[nocompress]{cite}
\else
  \usepackage{cite}
\fi

\usepackage{graphicx}
\ifCLASSINFOpdf
  
\else
 
\fi

\usepackage[rawfloats=true]{floatrow} 
\begin{document}
%

\title{FedBWO: Enhancing Communication Efficiency in Federated Learning}

\author{Vahideh Hayyolalam\IEEEauthorrefmark{1}, \textit{Student Member, IEEE}, Öznur Özkasap\IEEEauthorrefmark{1}, \textit{Senior Member, IEEE}\\ \IEEEauthorrefmark{1}Department of Computer Engineering, Koç University, İstanbul, Türkiye \\ Emails: \IEEEauthorrefmark{1}\{vhayyolalam20, oozkasap\}@ku.edu.tr}
%

%
%





\maketitle

\begin{abstract}
    Federated Learning (FL) is a distributed Machine Learning (ML) setup, where a shared model is collaboratively trained by various clients using their local datasets while keeping the data private. Considering resource-constrained devices, FL clients often suffer from restricted transmission capacity. Aiming to enhance the system performance, the communication between clients and server needs to be diminished. Current FL strategies transmit a tremendous amount of data (model weights) within the FL process, which needs a high communication bandwidth. Considering resource constraints,  increasing the number of clients and, consequently, the amount of data (model weights) can lead to a bottleneck. In this paper, we introduce the Federated Black Widow Optimization (FedBWO) technique to decrease the amount of transmitted data by transmitting only a performance score rather than the local model weights from clients. FedBWO employs the BWO algorithm to improve local model updates. The conducted experiments prove that FedBWO remarkably improves the performance of the global model and the communication efficiency of the overall system. According to the experimental outcomes, FedBWO enhances the global model accuracy by an average of 21\% over FedAvg, and 12\% over FedGWO. Furthermore, FedBWO dramatically decreases the communication cost compared to other methods.
\end{abstract}

%
\IEEEpeerreviewmaketitle

\textcolor{blue}{
}

\section{Introduction} \label{sec.intro}
    \IEEEPARstart{R}{ecently}, the widespread adoption of Internet of Things (IoT) devices, particularly smartphones and tablets, has led to a surge in diverse data generation in various data types, including images, text, audio, and sensor data, offering valuable opportunities for Machine Learning (ML) \cite{abdulrahman2020fedmccs}. However, collecting and storing large-scale data on centralized servers incurs high storage and network costs, which grow with data volume \cite{almanifi2023communication}.
    
     Federated Learning (FL) addresses these challenges by enabling local model training on devices, preserving privacy, and reducing data transfer. Despite this, FL faces issues like limited device computation, unstable networks, and high communication costs. Optimizing network transmission is essential to fully leverage FL for mobile data.

      Designing a robust ML model based on Artificial Neural Networks (ANNs) requires minimizing the volume of collected data, enhancing data security, and reducing the number of training parameters, such as the weights in the ANN. FL supports these goals by sharing only model updates, though communication time becomes a major bottleneck. In conventional centralized ML models, computation time typically outweighs communication time. Techniques such as Graphics Processing Units (GPUs) and parallel computing help mitigate this. However, in FL, the primary bottleneck shifts to network communication time, necessitating the optimization of network transmission. Most FL implementations use the Federated Averaging (FedAvg) algorithm \cite{mcmahan2017communication}, where clients compute updates locally and send them to a central server. While algorithms like FedAvg \cite{mcmahan2017communication} are common, they converge slowly in dynamic environments. Metaheuristic algorithms \cite{hayyolalam2017qos} such as Particle Swarm Optimization (PSO) \cite{park2021fedpso}, Grey Wolf Optimizer (GWO) \cite{abasi2022grey}, and Sine Cosine Algorithm (SCA) \cite{abasi2022sine} have been explored to accelerate convergence.

      This paper introduces the Federated Black Widow Optimization (FedBWO) framework, leveraging the BWO algorithm \cite{hayyolalam2020black} to improve FL efficiency by enhancing model performance and reducing communication costs. Inspired by black widow spiders' behavior, BWO effectively solves complex optimization problems. Our proposed FedBWO framework focuses on minimizing communication costs while maintaining high model accuracy, which optimizes model updates based on local accuracy or loss, avoiding large data transfers. Experimental results show FedBWO outperforms FedAvg \cite{mcmahan2017communication}, FedPSO \cite{park2021fedpso}, FedSCA \cite{abasi2022sine}, and FedGWO \cite{abasi2022grey}. in both model performance and communication efficiency. This research contributes to ongoing efforts to optimize FL in decentralized, resource-constrained environments while maintaining high model performance and minimizing communication costs.
    To the best of the authors’ knowledge, no existing research has effectively integrated the BWO algorithm into the FL framework to simultaneously address communication costs and model accuracy in mobile environments.

     This work makes several key contributions. First, it introduces the novel integration of the BWO algorithm into FL to accelerate model updates and enhance communication efficiency by leveraging BWO's unique exploration and exploitation capabilities for improved training performance. Second, the proposed FedBWO framework focuses on reducing communication costs by optimizing model parameters based on local accuracy and loss, minimizing data exchange with the server. Third, experimental results demonstrate that FedBWO outperforms existing methods like FedAvg \cite{mcmahan2017communication}, FedPSO \cite{park2021fedpso}, FedSCA \cite{abasi2022sine}, and FedGWO \cite{abasi2022grey} in both communication efficiency and model performance, highlighting its robustness and real-world applicability. Lastly, this research advances the field of FL by providing an effective optimization strategy that addresses challenges in decentralized, resource-constrained environments and offers valuable insights for future research.

    The rest of the paper is organized as follows. Section \ref{sec.rel} discusses the preliminaries and related work. The methodology of the proposed FedBWO is provided in section \ref{sec.metho}. The details of experiments are thoroughly discussed in section \ref{sec.exp}. Finally, section \ref{sec.conclusion} concluded the paper.

\section{Preliminaries and Related Work} \label{sec.rel}
\subsection{Federated Learning}
    Federated Learning (FL) is a distributed machine learning approach where clients collaboratively train a model without sharing their data. Unlike conventional machine learning, where data must be transferred to a central server for training, FL keeps data local, addressing security, privacy, and resource consumption issues. Traditional models fail to capture client-specific patterns, whereas FL allows personalized local training and reduces network traffic and latency \cite{banabilah2022federated}.

\begin{figure}[htbp!]
    \centering
    \includegraphics[width=1\linewidth]{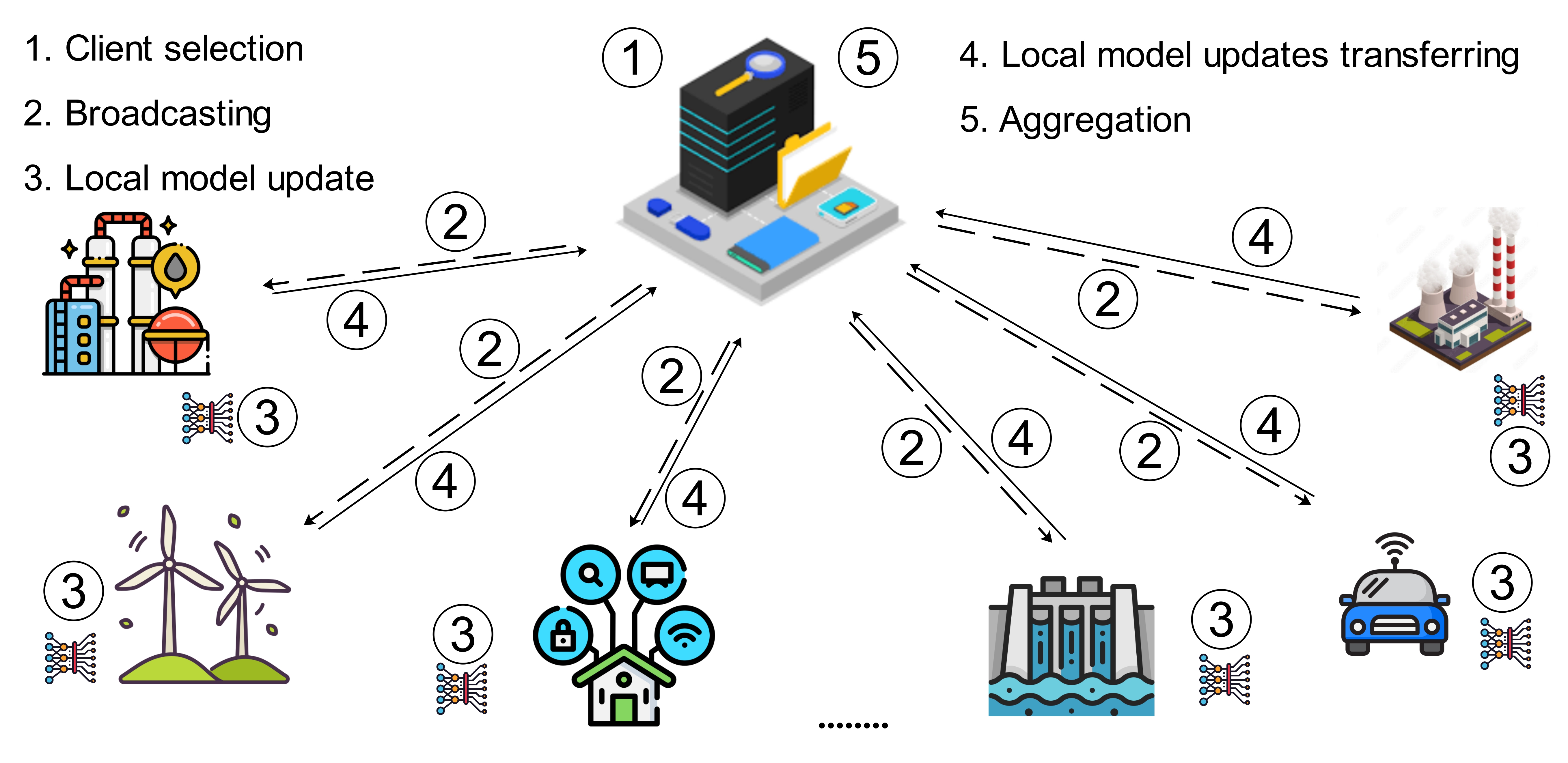}
    \caption{General Overview of Federated Learning}
    \label{fig.fl}
\end{figure}

    Vanilla Federated Learning (FL) involves two main components: clients (data owners) and the aggregator (server). Clients train models on their private data, while the server distributes the global model, aggregates local updates, and refines the model before sharing it back with clients \cite{pouriyeh2022secure}. The FL process begins by selecting clients for training, either randomly or based on specific criteria. The server then broadcasts the global model to the selected clients, who perform local training on their data. After training, clients send their updated model parameters to the server, which aggregates these updates, typically through averaging, to improve the global model. This cycle repeats until a stopping condition, such as a set number of iterations or a target performance level, is reached. Figure \ref{fig.fl} illustrates this process.

\subsection{Black Widow Optimization}

    The Black Widow Optimization (BWO) algorithm \cite{hayyolalam2020black} is a population-based meta-heuristic inspired by the mating and cannibalistic behaviors of black widow spiders (Algorithm \ref{alg.bw}). It begins with a random population and alternates between two phases: exploration and exploitation.

    In \textbf{exploration}, new solutions are generated through a mating process, expanding the search space. In \textbf{exploitation}, weaker offspring are eliminated, allowing stronger solutions to evolve. Mutations are applied to survivors to maintain diversity and avoid local optima.
    
    This balance makes BWO effective for optimization tasks like load balancing, resource allocation, and federated learning. The process repeats until a stopping condition, such as a set number of iterations or a fitness goal, is met.

\begin{algorithm}[htbp!] 
    \SetAlgoLined
    \small
        Initialize $N, MaxItr, Pm, Pc$. \\
        Create an initial population of $N$ solutions \\
        Evaluate the fitness of the population \\
        \For{$iteration=1, 2, ..., MaxItr$}{
           \textbf{Mating phase}: Generate offspring from parent pairs.\\
           \textbf{Combine} parents and offspring into one population. \\
           \textbf{Cannibalism phase}: Remove $Pc\%$ worst-performing individuals. \\
           \textbf{Mutation phase}: For each individual, with probability $Pm$, mutate the individual. \\
           \textbf{Sort} population by fitness. \\
           \textbf{Select} the best $N$ individuals for the next generation. \\
     }
      \textbf{return} the best solution.\\ 
     \caption{Pseudo-code of BWO (MaxItr: Maximum number of iterations)}
     \label{alg.bw}
    \end{algorithm}
\subsection{Related Work}
    Several studies in the literature employ meta-heuristic algorithms to enhance FL-based systems. According to the objective of adopting a meta-heuristic algorithm in the FL structure, we categorized the existing research works into four essential categories, including communication efficiency \cite{abasi2022grey, park2021fedpso, hossain2023fedavo, abasi2022sine}, security \& privacy enhancement \cite{khatua2024federated, vaiyapuri2023metaheuristics}, model enhancement \cite{kumbhare2023federated, agrawal2021genetic}, and resource optimization \cite{wen2024resource, yang2023fedrich}. 

    One of the most common applications of metaheuristic algorithms on FL is to enhance the communication efficiency of the whole FL process, which is our focus in this paper. Thus, we only investigate the studies with a particular concentration on communication efficiency enhancement via metaheuristic algorithms.

   Researchers in FedGWO \cite{abasi2022grey} have applied the Grey Wolf Optimizer to reduce communication costs by optimizing global model updates based on local loss and accuracy. The authors in FedPSO \cite{park2021fedpso} have used Particle Swarm Optimization to improve update efficiency, reducing server communication while maintaining accuracy, outperforming methods like FedAvg \cite{mcmahan2017communication}. Furthermore, researchers in FedSCA \cite{abasi2022sine} have adopted the Sine Cosine Algorithm, optimizing updates with performance metrics instead of weight transfers, achieving higher accuracy and lower communication costs in dynamic networks.

     The authors in FedAVO \cite{hossain2023fedavo} have employed the African Vulture Optimizer to automatically tune hyperparameters like client selection ratio and local training passes, reducing communication overhead and boosting global model accuracy by 6\% compared to existing methods (e.g., FedAvg \cite{mcmahan2017communication}, FedProx \cite{li2020federated}, FedPSO \cite{park2021fedpso} ).

\section{FedBWO Methodology} \label{sec.metho}

\subsection{Motivation}
    To enhance the accuracy of artificial neural networks (ANNs), one widely used technique is to increase the depth of the model by adding more layers, resulting in what is known as a deep neural network (DNN). As the number of layers grows, the quantity of weight parameters that need to be trained also rises significantly. This leads to a sharp increase in network communication costs when models are transferred from clients to servers in an FL setting, particularly in scenarios where large models are involved.

\begin{figure}[htbp!]
        \centering
        \includegraphics[width=1.1\linewidth]{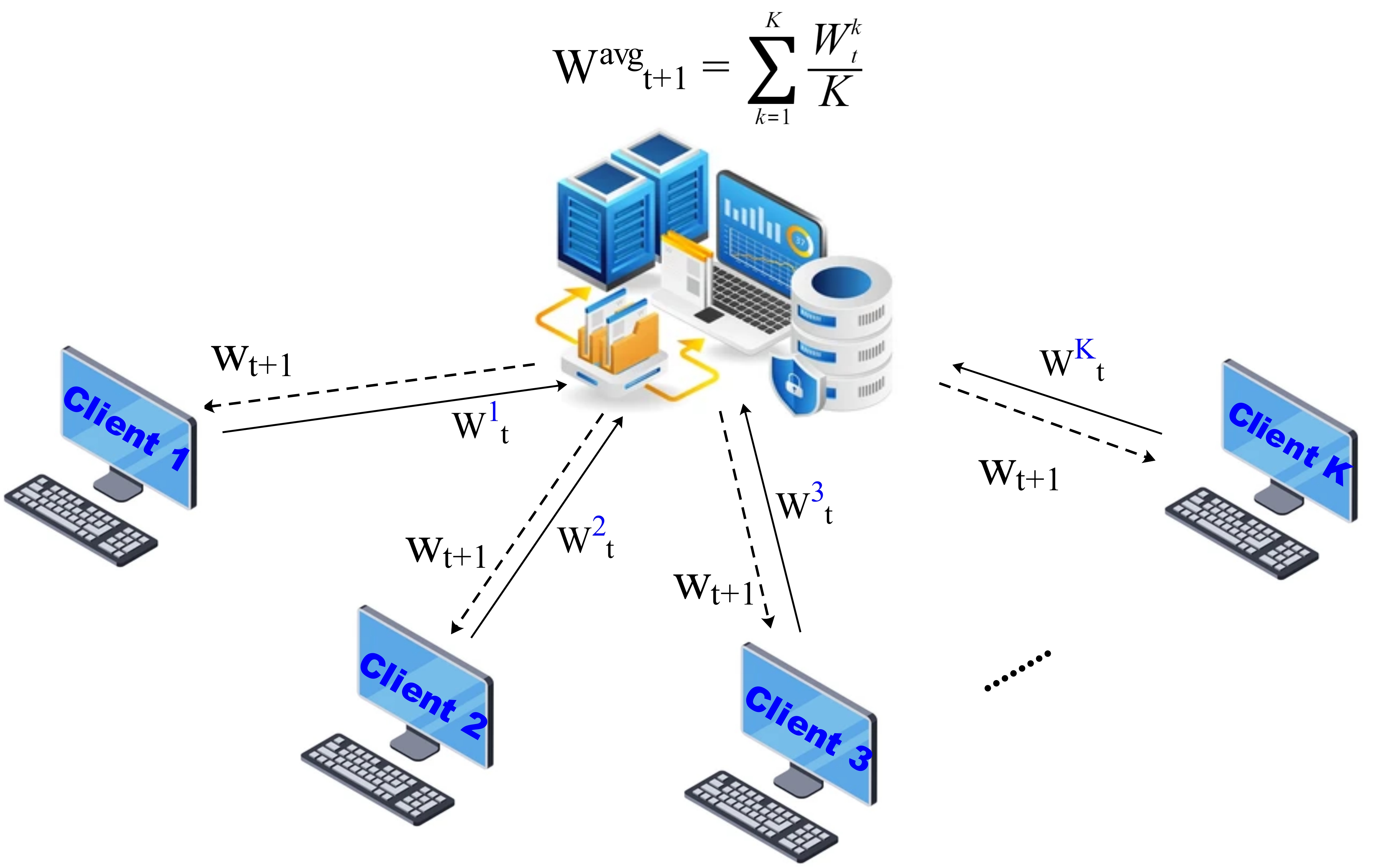}
        \caption{The weighted aggregation method (like FedAvg) calculates the average of the weight value obtained from the K clients and then transmits the updated weight ($W_{t + 1}$) back to the client.}
        \label{fig.avg}
\end{figure}

    In a typical FL setup (as illustrated in Figure \ref{fig.avg}), where client devices collaboratively train a shared model without exchanging raw data, the communication overhead can become a critical bottleneck. As the model's complexity increases, transferring the entire model to a central server becomes inefficient and costly. To address this, we propose a novel approach using the Black Widow Optimization (BWO) algorithm, called FedBWO. Unlike traditional methods, which send the entire model or focus solely on accuracy, our method transmits only the most relevant metrics, such as loss or accuracy, based on BWO’s selection criteria, thereby reducing the communication overhead. This approach enables the efficient transmission of optimized model parameters, regardless of the model's size, improving overall system performance in resource-constrained environments. 
    
    To set the stage for the FedBWO method, we will first review the algorithm employed in existing federated learning approaches, such as FedAvg. The procedure in Algorithm \ref{alg.fedAvg} for federated learning is as follows: first, clients participating in the training round are selected as outlined in Line 4. The weight values learned from these clients are collected according to Lines 5 and 6. Once the weights are collected, their average is calculated through Line 7, which is then used to update the global weights. The server sends these global weights to the clients, who use them to train their models, as described in lines 11-14.  

\begin{algorithm}[htbp!] 
    \SetAlgoLined
    \small
    \SetKwFunction{FS}{ServerRun}
    \SetKwProg{Fn}{Function}{:}{}
    \Fn{\FS{}}{
         Initialize $w^0$ \\
         \For{$t=0, 1, ..., T$ communication rounds}{
           Sc = (randomly select up to $max(C * K, 1)$ clients)\\
                \For{each client $k \in  $ Sc \textbf{in parallel}}{
                   $ w_{t+1}^k$ = \textbf{UpdateClient}($k$, $ w_{t}$)\\
                    $ w_{t+1}$ = aggregate ($ w_{t+1}^k$ )\\
                }
     }
    }
    \textbf{End Function}\\
    \SetKwFunction{CLo}{UpdateClient}
    \SetKwProg{Fn}{Function}{:}{}
    \Fn{\CLo{$k$, $ w$}}{
          Execute training on client $k$ with $w$ for $E$ epochs  \\ 
          $w$ = updated weight after training\\       
        \textbf{return} $w$ \\  
    }
    \textbf{End Function}\\ 
     \caption{FedAvg algorithm; $K$ = number of clients; $E$ = client total epochs; Select client by the $C$ ratio \cite{park2021fedpso}.}
     \label{alg.fedAvg}
    \end{algorithm}
\subsection{Proposed Method}
\begin{figure}[htbp!]
        \centering
        \includegraphics[width=1\linewidth]{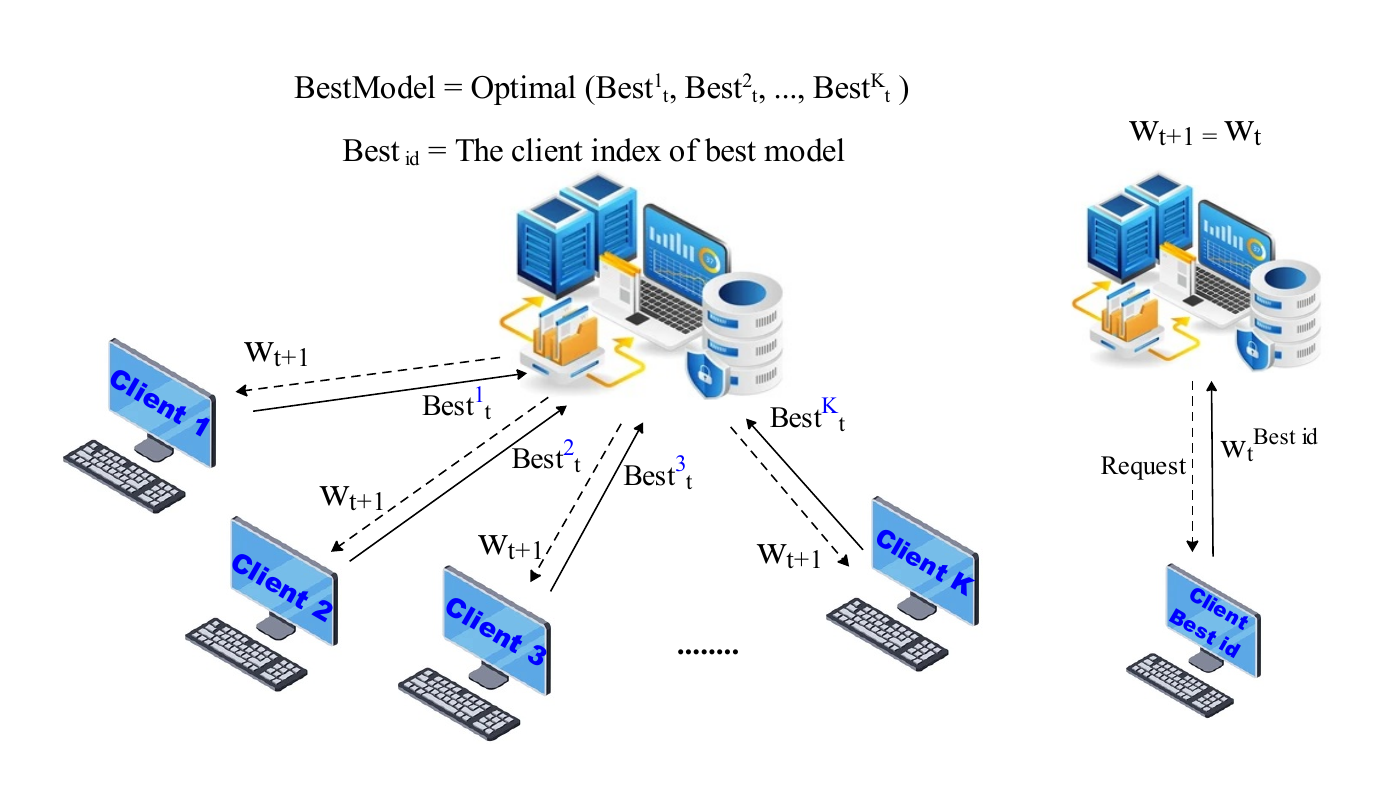}
        \caption{The procedure of updating the weights in the FedBWO algorithm (The server requests the client with the highest score to serve as the global model after collecting scores from all clients.} 
        \label{fig.FedBWO} 
\end{figure}

    In the proposed FedBWO model, only the weights from the client with the best performance score are sent to the server. Consequently, there is no need to transmit model weights from every client. This procedure is illustrated in Figure \ref{fig.FedBWO}. 

    The best score is determined by the lowest loss value achieved after the training process on the clients, which is typically only four bytes in size. FedBWO leverages the features of the BWO algorithm to identify the top-performing model and updates the global model based on this information. The weight update for FedBWO can be described using the principles of BWO:
    \begin{itemize}
        \item Selection: Choose the best-performing solutions (weights) based on evaluation (e.g., lowest loss).
        \item Procreate: Combine parts of the best solutions to create new weights.
        \item Cannibalism: Simulate the cannibalistic behavior where weaker solutions are eliminated.
        \item Mutation: Introduce random changes to some of the offspring to maintain diversity.
    \end{itemize}

    Based on these principles, we introduce the FedBWO algorithm in Algorithm \ref{alg.fedBwo}, which builds upon Algorithm \ref{alg.fedAvg} by integrating BWO. Unlike traditional methods, the \textbf{ServerRun} function receives only the best scores instead of the weights (w) from the clients on Line 5. Identifying the client with the lowest score among those collected is handled in Lines 7–10. The \textbf{UpdateClient} function applies the BWO to the ANN, encompassing selection, procreation, cannibalism, and mutation steps for updating the weights. This process is repeated for each layer's weights. The best scores from each client are then sent to the server. The \textbf{GetBestModel} function is used to request the model from the client with the highest score on the server.

\begin{algorithm}[htbp!] 
    \SetAlgoLined
    \SetKwFunction{FS}{ServerRun}
    \SetKwProg{Fn}{Function}{:}{End Function}
    \Fn{\FS{}}{
         Initialize $w^0, Best, Best_{id},$ and parameters of BWO \\
         \For{$t=0, 1, ..., T$ communication rounds}{
                \For{each client $k$ \textbf{in parallel}}{
                   $Best^k$ = \textbf{UpdateClient}($ w_{t}$)\\
                
                \If{$Best^k$ is better than $Best $}
                {
                $Best = Best^k$\\
                $Best_{id} = k $
                }
                }
     }
     BestModel = \textbf{GetBestModel}($Best_{id}$)
    }
\SetKwFunction{CLo}{UpdateClient}
    \SetKwProg{Fn}{Function}{:}{End Function}
    \Fn{\CLo{$ w$}}{
          \For{each weight layer $l$ }{
                   Update the weight using BWO operations
                }      
        \textbf{return} $Best^k$ \\  
    }
%
 \SetKwFunction{GB}{GetBestModel}
    \SetKwProg{Fn}{Function}{:}{End Function}
    \Fn{\GB{$Best_{id}$}}{
          Request from server to client($Best_{id}$)\\
          Transfer wights from client ($Best_{id}$) to server\\
          \textbf{return} BestModel \\  
    }
%
     \caption{FedBWO}
     \label{alg.fedBwo}
    \end{algorithm}

\subsection{The Solution Representation}
    In the context of using the BWO algorithm for federated learning, the solution representation involves encoding the weights of ANN as potential solutions. In the FedBWO framework, the BWO algorithm is adapted to optimize ANN weights for FL. Unlike the original operation order of BWO, in FedBWO, the process begins with mutation, introducing random changes to the existing solution to enhance diversity and explore new areas of the solution space. This is followed by procreation, where segments of the solutions are combined to generate new solutions. Finally, cannibalism is applied to eliminate weaker solutions, retaining only the most optimal one. After this process, clients send their fitness scores to the server, which selects the best-performing client to share its optimized weights. The server then updates the global model accordingly.

\section{Experimental Evaluation} \label{sec.exp}
    Aiming to examine the efficiency and effectiveness of the proposed FedBWO, we conducted experiments to evaluate the model performance and communication cost.
    We assessed and analyzed to determine whether the proposed model achieves reasonable performance in terms of accuracy and loss, along with communication cost, given its lower network connection requirements than FedAvg. We have adopted the CIFAR-10 dataset as our experimental data to analyze and compare the test accuracy of the proposed FedBWO against FedPSO \cite{park2021fedpso}, FedGWO \cite{abasi2022grey}, FedSCA \cite{abasi2022sine}, and FedAvg \cite{mcmahan2017communication}. Also, we reviewed the data communication cost between clients and the server. To ensure the validity of the experiment, all results were obtained by averaging the outcomes over 10 separate runs.
    
\subsection{Experimental Setup}
    The experiments are conducted using High-Performance Computing (HPC) with TensorFlow and Keras for implementation. The proposed FedBWO aims to enhance the communication performance of FL by updating model weights through the BWO algorithm, reducing data transferred between clients and the server. The experiments utilize a two-layered CNN model, where the first layer has 32 channels and the second has 64 channels, each followed by a $2 \times 2$ max pooling layer. This CNN architecture is adopted from FedAvg \cite{mcmahan2017communication}, FedPSO \cite{park2021fedpso}, FedGWO \cite{abasi2022grey}, and FedSCA \cite{abasi2022sine}. Specifically, the model consists of a Conv2D layer with a $5 \times 5 \times 32$ kernel, followed by another Conv2D layer with 32 filters. The second convolutional block includes a Conv2D layer with a $5 \times 5 \times 64$ kernel and another Conv2D layer with 64 filters. This is followed by a Dense layer with $1024 \times 512$ units, another Dense layer with 512 units, and finally, a Dense layer with $512 \times 10$ units leading to the output layer with 10 units.

    We have employed the CIFAR-10 dataset for conducting the experiments. CIFAR-10 is a popular image dataset that is widely used for image classification tasks. This dataset encompasses 50,000 training and 10,000 testing 32 x 32-pixel images, belonging to ten various categories, such as cats, automobiles, and airplanes. The CIFAR-10 dataset was shuffled, assigned to client numbers, and distributed accordingly to proceed with training. No specific tuning procedures were applied to enhance accuracy during the training phase, apart from adjustments to the dropout layer. All the experimental models, as well as FedBWO, utilized SGD for client training, with the learning rate set at 0.0025. The specific hyperparameter values adopted in the experiments are the same for all experimental models. The number of clients, client epoch, Batch size, and Epoch are set to 10, 5, 10, and 30, respectively. 
    
\subsection{Accuracy Evaluation}
\begin{figure}[htbp!]
        \centering
        \includegraphics[width=1\linewidth]{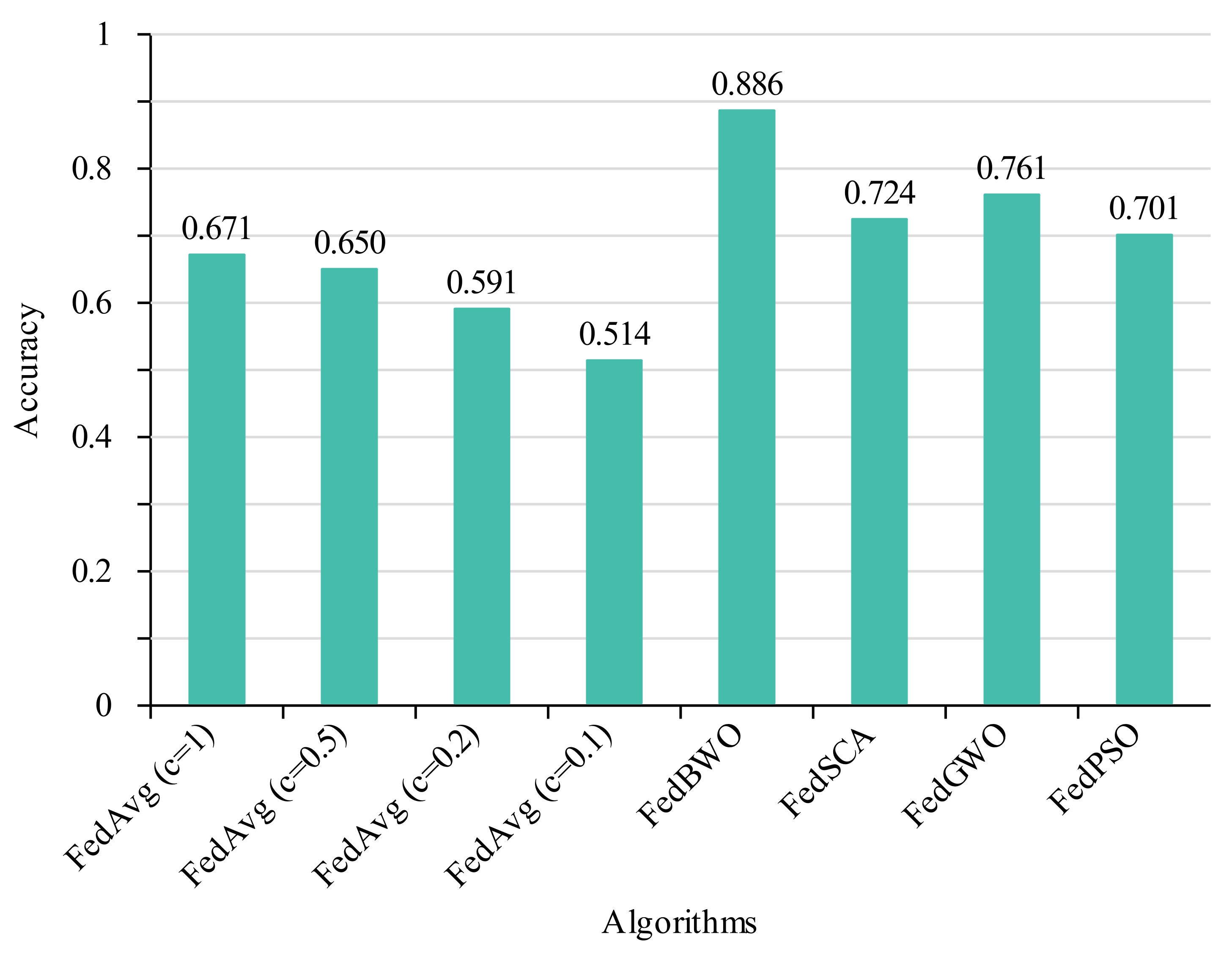}
        \caption{Accuracy comparison.} 
        \label{fig.avg2}
\end{figure}

    The experimental results shown in Figure \ref{fig.avg2} reveal that FedBWO achieved the highest accuracy of 88.64\%, significantly outperforming other FL algorithms. This improvement is due to the effective exploration and exploitation capabilities of the BWO algorithm, which significantly optimizes model updates. In comparison, FedAvg achieved lower accuracy, ranging from 51.39\% with $C = 0.1$ to 67.14\% with $C = 1.0$. Metaheuristic-based methods like FedPSO (70.13\%), FedSCA (72.42\%), and FedGWO (76.10\%) performed better than FedAvg but still were outperformed by FedBWO. FedBWO’s superior accuracy highlights its ability to balance exploration and exploitation, leading to a more efficient FL environment.

\subsection{Loss Evaluation}
\begin{figure}[htbp!]
        \centering
        \includegraphics[width=1\linewidth]{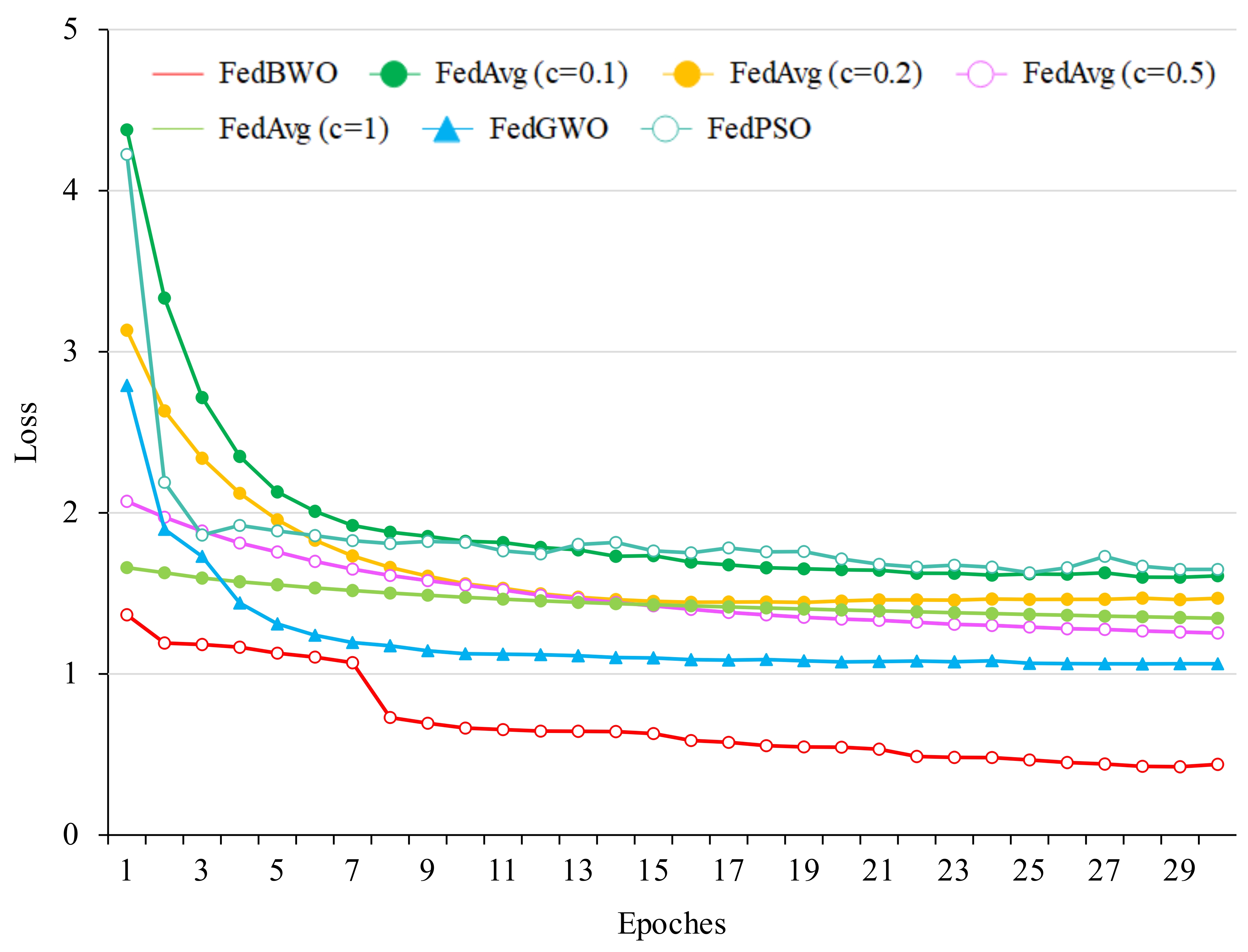}
        \caption{Loss comparison.} 
        \label{fig.loss}
\end{figure}

    The experimental results illustrated in Figure \ref{fig.loss} indicate that FedBWO has achieved the lowest loss among all compared algorithms, demonstrating its superior optimization capability. This improvement is attributed to the BWO algorithm's effective balance between exploration and exploitation, which ensures more efficient model updates

\subsection{Communication Cost Evaluation}
    The communication cost in FL is significantly influenced by client participation and data volume. This study analyzes these factors, explains communication cost calculation, and compares the proposed FedBWO with other methods for communication efficiency.

    The \textbf{Client Selection Parameter} ($C$), ranging from 0 to 1, determines the fraction of clients participating in each round. A higher $C$ increases client participation and data transmission, raising communication costs. For example, $C = 1.0$ means all clients participate, while $C = 0.5$ includes only half. The \textbf{data transferred per round} depends on $C$ and the total number of clients ($N$), calculated as $C \times N \times D$, where $D$ is the amount of transmitted data during each round from a client to the server.

    Communication cost differs between methods. In FedAvg, all selected clients transmit their model weights, with the cost per round calculated as $C \times N \times M$, where $M$ is the model size in bytes. In contrast, algorithms like FedBWO, FedPSO, FedGWO, and FedSCA only transmit $4-byte$ scores from all clients, requesting full model weights only from the best-performing client. Their cost per round is $N \times 4 + M + \epsilon$, where $\epsilon$ (the server request) is negligible. This design significantly reduces communication compared to FedAvg.

    The total communication cost over $T$ rounds is given by Equations \ref{eq.ave} and \ref{eq.x} for FedAvg and $FedX$ (Algorithms like FedBWO, FedPSO, FedGWO, and FedSCA), respectively.
      \begin{equation} \label{eq.ave}
         Total Cost_{FedAvg} = T \times C \times N \times M
      \end{equation}    
      \begin{equation} \label{eq.x}
         Total Cost_{FedX} = T \times (N \times 4 + M + \epsilon)
      \end{equation}
  To fairly compare these methods, communication cost is normalized by Equation \ref{equ.n}, where we considered $C=1.0$. Assuming the number of communication rounds is constant, the term $T$ can be omitted from the equation. Under this assumption, the communication cost for all $FedX$ algorithms becomes identical and remains constant, approximately equal to FedAvg when $C=0.1$, since the total number of clients is constant, $N = 10$ and by considering $C=0.1$, only one client will be selected.
  \begin{equation} \label{equ.n}
     Normalized Cost_{FedX} = \frac{T \times (N \times 4 + M + \epsilon)}{T \times C \times N \times M}
  \end{equation}
   Keeping the above mentioned in mind, to fairly compare $FedX$ algorithms against each other, we considered stop conditions for communication rounds. The iterative process is terminated under one of the following conditions:
    \begin{itemize}
        \item There are no significant changes in performance for $t$ consecutive iterations.
        \item The accuracy reaches above a pre-defined threshold ($\tau$).
        \item The number of communication rounds reaches a predefined limit of the number of global epochs.
    \end{itemize}

    As previously mentioned, the total number of clients is set to $N = 10$ in all the experiments, and in Equation \ref{equ.n}, $C = 1.0$ is used. The numerator of Equation \ref{equ.n} simplifies to $T \times M$ since the term $10 \times 4 + \epsilon$ is negligible. Consequently, this simplification results in Equation \ref{equ.n4}, where $T_{X}$ and $T_{Ave}$ represent the number of epochs completed by the respective FedX algorithm and FedAvg ($C = 1.0$), respectively.




     \begin{equation} \label{equ.n4}
     Normalized Cost_{FedX} = \frac{T_{X}}{T_{Ave} \times 10}
    \end{equation}

\begin{figure}[htbp!]
        \centering
        \includegraphics[width=1\linewidth]{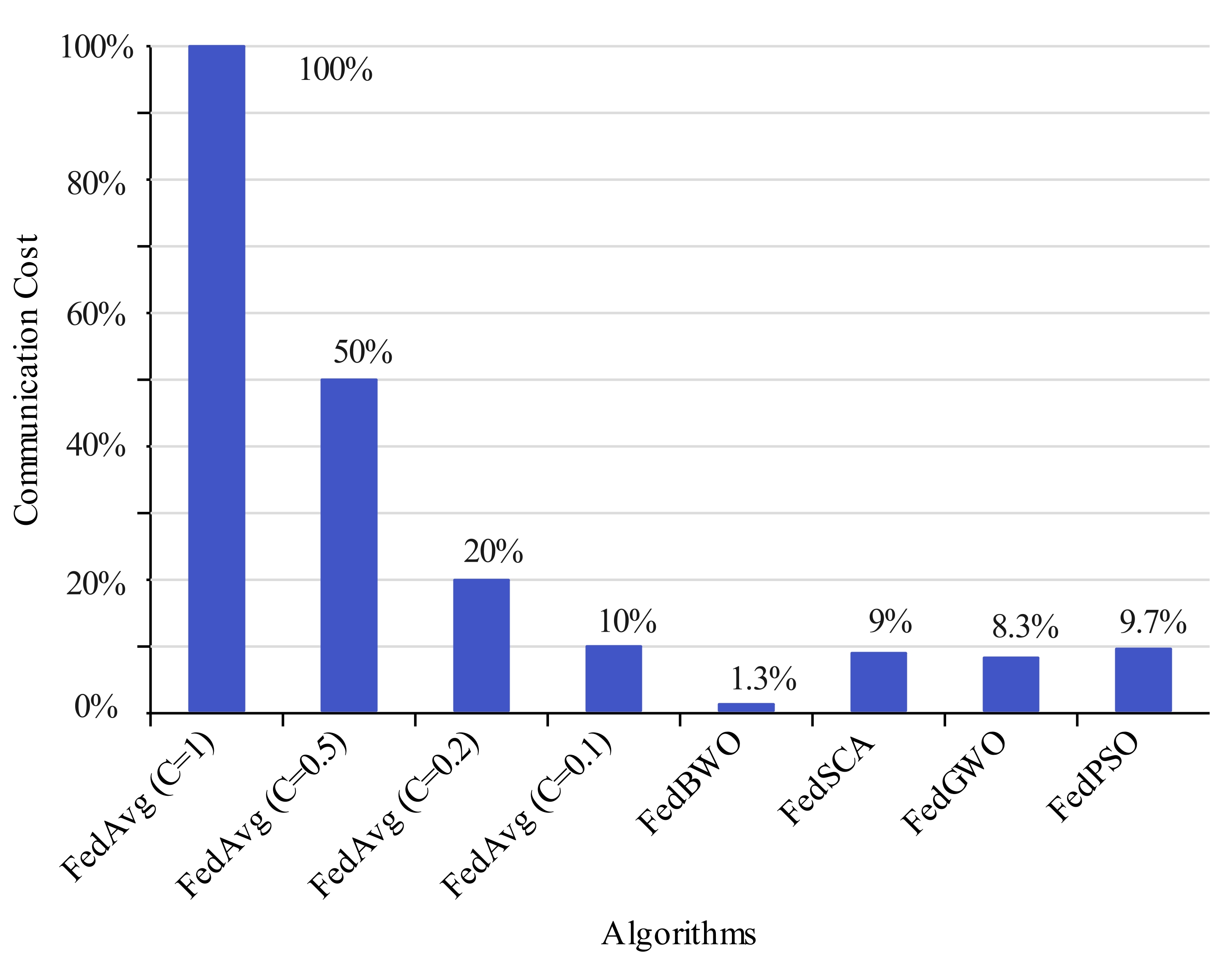}
        \caption{Communication Cost Comparison.} 
        \label{fig.comcost}
\end{figure}

    According to our experiments, where we set $t=5$ and $\tau = 70$, the total communication rounds completed by each algorithm highlight FedBWO's superior communication efficiency. FedBWO completed only 4 communication rounds, resulting in a 1.3\% communication cost relative to FedAvg with $C = 1.0$, which required 30 rounds. In comparison, FedPSO completed 29 rounds (9.7\% cost), FedSCA completed 27 rounds (9\% cost), and FedGWO completed 25 rounds (8.3\% cost). Additionally, variations of FedAvg with different client participation rates showed reduced costs: 50\% for $C = 0.5$, 20\% for $C = 0.2$, and 10\% for $C = 0.1$, all completing 30 rounds. Figure \ref{fig.comcost} illustrates the communication cost evaluation across these FL algorithms. FedBWO's early convergence and minimal data transmission make it the most communication-efficient approach in this evaluation.

\subsection{Execution Time Evaluation}
\begin{figure}[htbp!]
        \centering
        \includegraphics[width=1\linewidth]{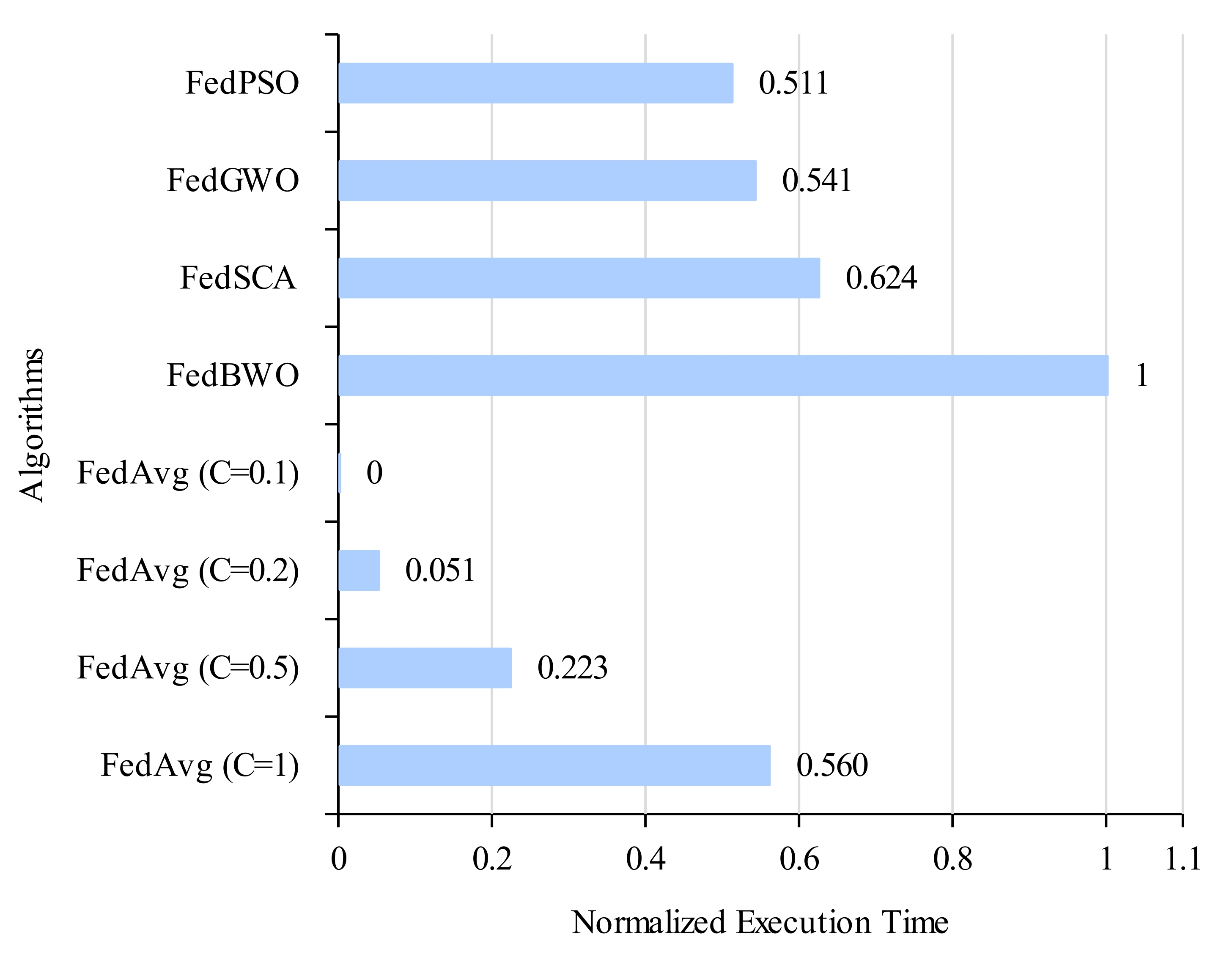}
        \caption{Execution Time Comparison.}
        \label{fig.time}
    \end{figure}

    To evaluate the execution time of our experimental approaches, we first measured their execution time, then, normalized the observed values in between $0$ and $1$. Figure \ref{fig.time} illustrates the obtained result for this experiment. The result indicates that FedBWO remarkably has a high execution time, which makes it unsuitable for real-time applications.

\section{Conclusion} \label{sec.conclusion}
    This study introduced the FedBWO framework, a novel approach designed to address communication bottlenecks in FL, particularly in resource-constrained environments. By leveraging the BWO algorithm, FedBWO reduces communication overhead by transmitting only $4-byte$ performance scores from clients instead of full model weights. This strategy significantly decreases data transmission without compromising model performance. Extensive experiments conducted on the CIFAR-10 dataset demonstrated FedBWO's superior performance in both communication efficiency and global model performance. However, its higher execution time makes it less suitable for real-time applications compared to alternatives like FedAvg and FedPSO. Overall, FedBWO effectively balances communication efficiency and model performance, making it a promising solution for FL systems with limited network capacity. Future work will focus on reducing execution time to expand FedBWO's applicability to real-time and latency-sensitive environments.


\ifCLASSOPTIONcompsoc
\else
\fi

\section*{Acknowledgment}
This work is supported by TÜBITAK (The Scientific and Technical Research Council of Türkiye) 2247-A National Outstanding Researchers Program Award 121C338.



%


 \input{FedBWOppr.bbl}

\end{document}